\documentclass{article}

\PassOptionsToPackage{numbers, compress}{natbib}
\usepackage[final]{nips_2017}

\usepackage[utf8]{inputenc} 
\usepackage[T1]{fontenc}    
\usepackage{hyperref}       
\usepackage{url}            
\usepackage{booktabs}       
\usepackage{amsfonts}       
\usepackage{nicefrac}       
\usepackage{microtype}      
\usepackage{graphicx}
\usepackage{subcaption}
\usepackage{amsmath}
\usepackage{bbm}
\usepackage{algorithm}
\usepackage[noend]{algpseudocode}
\usepackage[export]{adjustbox}
\usepackage{float}
\usepackage{bm}
\usepackage[multiple]{footmisc}
\usepackage{wrapfig}

\usepackage[colorinlistoftodos]{todonotes}

\title{Improving Factor-Based Quantitative Investing by 
Forecasting Company Fundamentals}

\author{
  John Alberg 
\\
  Euclidean Technologies\\
  \texttt{john.alberg@euclidean.com} \\
  \And
    Zachary C. Lipton 
    \\
	Amazon AI\\
    Carnegie Mellon University\\
  \texttt{zlipton@cmu.edu} \\
}

\begin{document}

\maketitle

\begin{abstract}
On a periodic basis,
publicly traded companies are required to report \emph{fundamentals}: financial data such as revenue, operating income, debt, among others.
These data points provide some insight into the financial health of a company.
Academic research has identified some factors, 
i.e. computed features of the reported data,
that are known through retrospective analysis 
to outperform the market average. 
Two popular factors are the book value normalized by market capitalization (book-to-market)
and the operating income normalized by the enterprise value (EBIT/EV).
In this paper, we first show through simulation that if we could (clairvoyantly)
select stocks using factors
calculated on \emph{future} fundamentals (via oracle), 
then our portfolios would far outperform 
a standard factor approach.
Motivated by this analysis,
we train deep neural networks to forecast future fundamentals 
based on a trailing 5-years window.
Quantitative analysis demonstrates a significant improvement in MSE over a naive strategy.
Moreover, in retrospective analysis using an industry-grade stock portfolio simulator (backtester),
we show an improvement in compounded annual return to 17.1\% (MLP) 
vs 14.4\% for a standard factor model.

\end{abstract}

\section{Introduction}
\label{sec:intro}
Public stock markets provide a venue 
for buying and selling shares, 
which represent fractional ownership 
of individual companies. 
Prices fluctuate frequently, 
but the myriad drivers of price movements 
occur on multiple time scales.  
In the short run, 
price movements might reflect 
the dynamics of order execution,  
and the behavior of high frequency traders.
On the scale of days, 
price fluctuation might be driven by the news cycle.
Individual stocks may rise or fall on rumors or reports of sales numbers, product launches, etc.
In the long run,
we expect a company's market value 
to reflect its financial performance,  
as captured in \emph{fundamental data}, i.e., 
reported financial information such as income, revenue, assets, dividends, and debt.
In other words, shares reflect ownership in a company 
thus share prices should ultimately move towards the company's \emph{intrinsic value}, 
the cumulative discounted cash flows associated with that ownership.
One popular strategy called \emph{value investing} is predicated on the idea that long-run prices reflect this intrinsic value 
and that 
the best features for predicting 
\emph{long-term} intrinsic value 
are the \emph{currently} available fundamental data.

In a typical quantitative (systematic) investing strategy, 
we sort the set of available stocks according to some \emph{factor} 
and construct investment portfolios comprised of those stocks 
which score highest.
Many quantitative investors engineer \emph{value factors} by taking fundamental data 
in a ratio to stock’s price, 
such as EBIT/EV or book-to-market. 
Stocks with high value factor ratios are called \emph{value} stocks and those with low ratios are called \emph{growth} stocks.
Academic researchers have demonstrated empirically that portfolios of stocks which overweight value stocks have significantly outperformed portfolios 
that overweight growth stocks over the long run
\citep{Leivo2017value, fama1992returns}.  

\begin{wrapfigure}{r}{0.5\textwidth}
\begin{center}
\includegraphics[width=\linewidth]{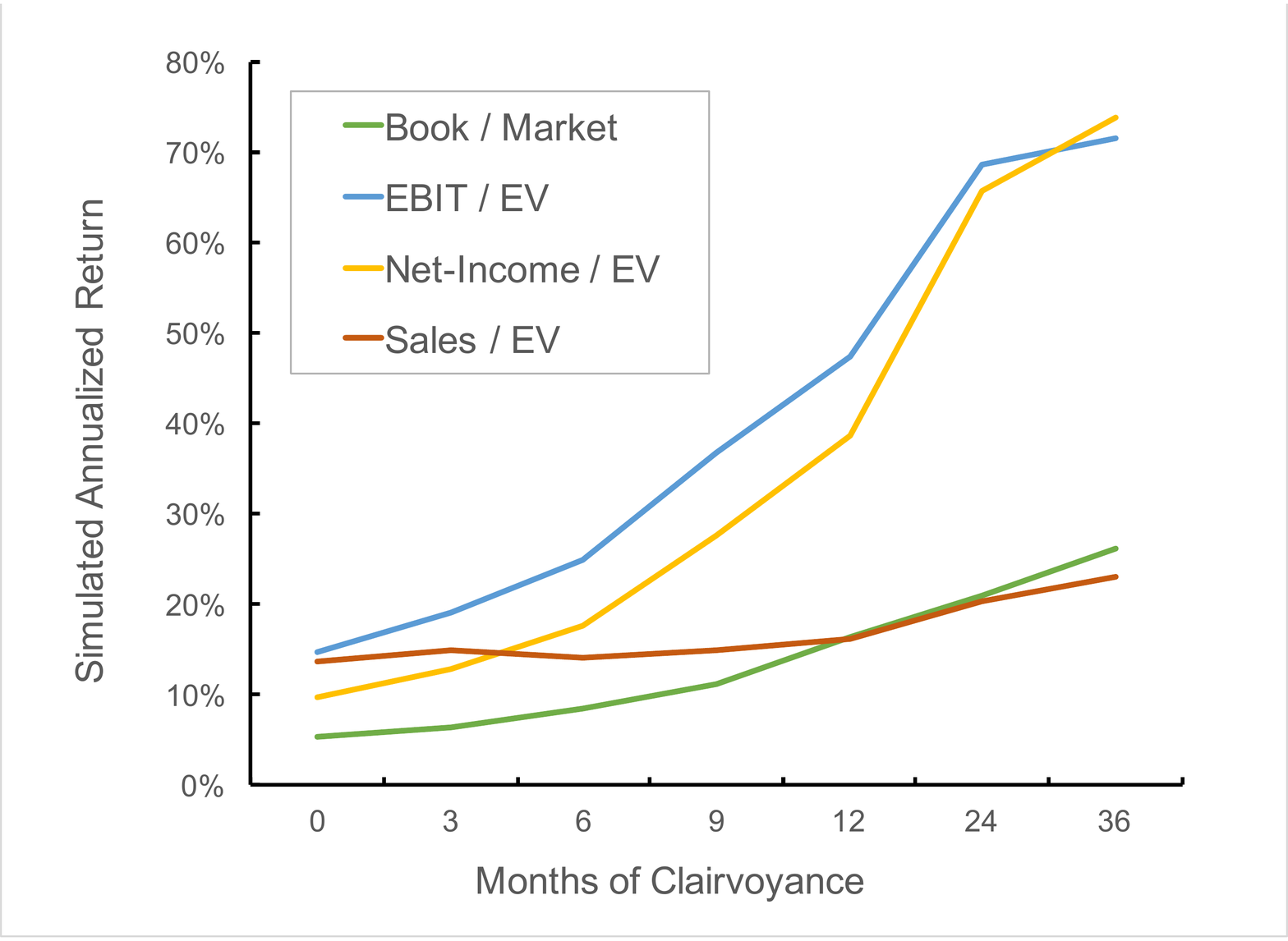}
\caption{Annualized return for various factor models for different degrees of clairvoyance.}
\label{fig:clairvoyant-factor-models}
\end{center}
\end{wrapfigure}

\textbf{In this paper, }
we propose an investment strategy 
that constructs portfolios of stocks today
based on \emph{predicted future fundamentals}.
Recall that value factors should identify companies 
that are inexpensively priced with respect to current company fundamentals such as earnings or book-value. 
We suggest that the long-term success of an investment 
should depend on the how well-priced the stock \emph{currently is} with respect to its \emph{future fundamentals}.
We run simulations with a \emph{clairvoyant model} 
that can access future financial reports (by oracle).
In Figure \ref{fig:clairvoyant-factor-models},
we demonstrate
that for the 2000-2016 time period,
a clairvoyant model applying the EBIT/EV factor with $12$-month clairvoyant fundamentals, if possible,
would achieve a $44$\% compound annualized return.

Motivated by the performance of factors applied to \emph{clairvoyant} future data, we propose to \emph{predict} future fundamental data 
based on trailing time series of 5 years of fundamental data.
We denote these algorithms as Lookahead Factor Models (LFMs). 
Both multilayer perceptrons (MLPs) and recurrent neural networks (RNNs) can make informative predictions,
achieving out-of-sample MSE of $.47$,
vs $.53$ for linear regression and $.62$ for a naive predictor.
Simulations demonstrate that investing with LFMs based on the predicted factors yields a compound annualized return (CAR) of $17.1$\%, vs 14.4\% for a normal factor model and a Sharpe ratio $.68$ vs $.55$.

\paragraph{Related Work}
\label{sec:related}
Deep neural networks models have proven powerful
for tasks as diverse as language translations \citep{sutskever2014sequence,bahdanau2014neural}, video captioning \citep{mao2014deep,vinyals2015show}, video recognition \citep{donahue2015long, tripathi2016context}, and time series modeling \citep{lipton2016learning,lipton2016directly, che2016recurrent}.
A number of recent papers consider deep learning approaches 
to predicting stock market performance. 
\cite{batres2015deep} evaluates MLPs for stock market prediction. 
\cite{ding2015deep} uses recursive tensor nets to extract events from CNN news reports
and uses convolutional neural nets 
to predict future performance 
from a sequence of extracted events. 
Several preprinted drafts consider deep learning for stock market prediction \cite{chen2015lstm,wanjawa2014ann,jia2016investigation} however, in all cases, 
the empirical studies are  limited to few stocks and short time periods.

\section{Deep Learning for Forecasting Fundamentals}
\label{sec:methods}
\paragraph{Data} In this research, we consider all stocks that were publicly traded on the NYSE, NASDAQ or AMEX exchanges for at least $12$ consecutive months between between January, 1970 and September, 2017.
From this list, we exclude non-US-based companies, 
financial sector companies, and any company with an inflation-adjusted market capitalization value below  
$100$ million dollars.
The final list contains $11,815$ stocks. 
Our features consist of reported financial information 
as archived by the \emph{Compustat North America} and \emph{Compustat Snapshot} databases. 
Because reported information arrive intermittently throughout a financial period,
we discretize the raw data to a monthly time step.
Because we are interested in long-term predictions
and to smooth out seasonality in the data,
at every month, we feed in inputs with a $1$-year lag between time frames and predict the fundamentals $12$ months into the future.

For each stock and at each time step $t$, 
we consider a total of $20$ input features.
We engineer $16$ features from the fundamentals as inputs to our models.
Income statement features are cumulative \emph{trailing twelve months}, denoted TTM, and balance sheet features are most recent quarter, denoted MRQ.
First we consider
These items include \emph{revenue} (TTM); 
cost of goods sold (TTM);
selling, general \& and admin expense (TTM);
earnings before interest and taxes or EBIT (TTM);
net income (TTM);
cash and cash equivalents (MRQ);
receivables (MRQ);
inventories (MRQ);
other current assets (MRQ);
property plant and equipment (MRQ);
other assets (MRQ);
debt in current liabilities (MRQ);
accounts payable (MRQ);
taxes payable (MRQ);
other current liabilities (MRQ);
total liabilities (MRQ).
For all features, we deal with missing values 
by filling forward previously observed values,
following the methods of \cite{lipton2016learning}.
Additionally we incorporate $4$ \emph{momentum features},
which indicate the price movement of the stock over the previous $1$, $3$, $6$, and $9$ months respectively. 
So that our model picks up on relative changes
and doesn't focus overly on trends in specific time periods,
we use the percentile among all stocks as a feature (vs absolute numbers). 

\textbf{Preprocessing \quad}
Each of the fundamental features exhibits 
a wide dynamic range over the universe of considered stocks. 
For example, Apple's 52-week revenue as of September 2016
was \$215 billion (USD). 
By contrast, National Presto, which manufactures pressure cookers, 
had a revenue \$340 million.
Intuitively, these statistics are more meaningful
when scaled by some measure of a company's size. 
In preprocessing, we scale all fundamental features in given time series by the market capitalization in the last input time-step of the series. We scale all time steps by the same value so that the neural network can assess the relative change in fundamental values between time steps. 
While other notions of size are used, such as enterprise value and book equity, we choose to avoid these measure 
because they can, although rarely, take negative values. 
We then further scale the features so that they each individually have zero mean and unit standard deviation.

\textbf{Modeling \quad}
In our experiments, we divide the timeline in to an \emph{in-sample} and \emph{out-of-sample} period. 
Then, even within the in-sample period, we need to partition some of the data as a validation set.
In forecasting problems, we face distinct challenges in guarding against overfitting. 
First, we're concerned with the traditional form of overfitting.
Within the in-sample period,
we do not want to over-fit to the finite observed training sample.
To protect against and quantify this form of overfitting, 
we randomly hold out a validation set consisting of $30\%$ of all stocks.
On this \emph{in-sample} validation set, we determine all hyperparameters,
such as learning rate, model architecture, objective function weighting.
We also use the in-sample validation set to determine early stopping criteria. 
When training, we record the validation set accuracy after each training epoch, 
saving the model for each best score achieved.
When $25$ epochs have passed without improving on the best validation set performance,
we halt training and selecting the model with the best validation performance. 
In addition to generalizing well to the in-sample holdout set,
we evaluate whether the model can predict the future \emph{out-of-sample} stock performance. Since this research is focused on long-term investing, we chose large in-sample and out-of-sample periods of the years 1970-1999 and 2000-2016, respectively.

\begin{figure}[t]
  \begin{subfigure}[b]{0.48\textwidth}
    \centering
  \begin{tabular}{l|ccc}
  \toprule
  Strategy & MSE & CAR & Sharpe Ratio\\
  \midrule
  {S\&P 500}&n/a&$4.5$\% & 0.19 \\
  {Market Avg.}& n/a & $7.7$\% & 0.29 \\
  {Price-LSTM}& n/a & $11.3$\% & 0.60 \\
  {QFM}& 0.62 & $14.4$\% & 0.55 \\
  {LFM-Linear}& 0.53 & $15.9$\% & 0.63 \\
  {LFM-MLP}& 0.47 & $17.1$\% & 0.68 \\
  {LFM-LSTM}& 0.47 & $16.7$\% & 0.67 \\
  \bottomrule
  \end{tabular}    
\caption{Out-of-sample performance for the 2000-2016 time period. All factor models use EBIT/EV. 
QFM uses current EBIT while our proposed LFMs use predicted EBIT. Price-LSTM is trained to predict relative return.
}
    \label{tab:simulation-results}
  \end{subfigure}
  \hfill
  \begin{subfigure}[b]{.48\textwidth}
      \centering
      \includegraphics[width=.8 \linewidth]{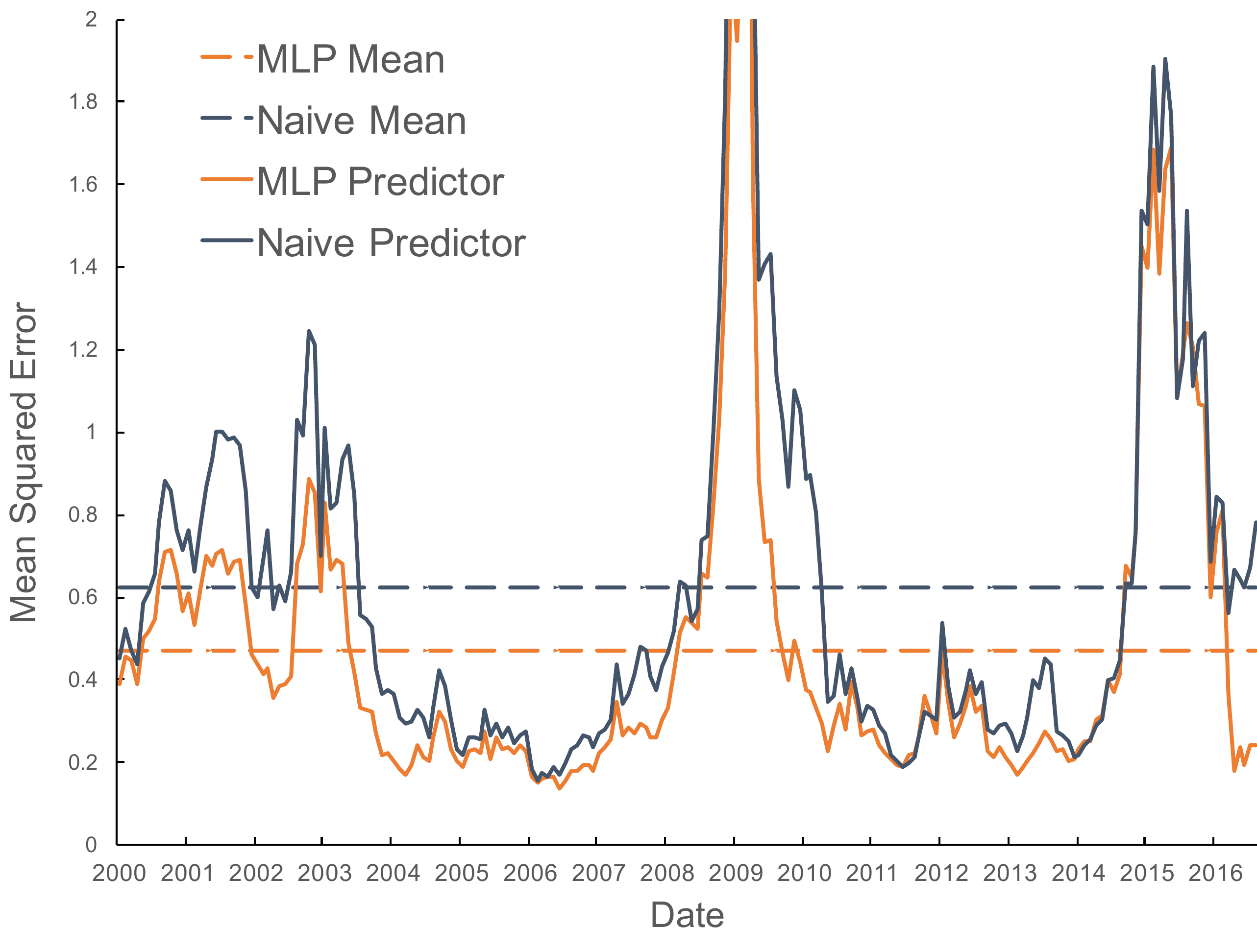}
      \caption{MSE over out-of-sample period for MLP (orange) and naive predictor (black).}
    \label{fig:out-of-sample-mse-mlp}
\end{subfigure}
\caption{Quantitative results}
\vspace{-30px}
\end{figure}

In previous experiments, we tried predicting relative returns directly with RNNs and while the RNN outperformed other approaches on the in-sample period, it failed to meaningfully out-perform a linear model (See results in Table  \ref{tab:simulation-results}).
Given only returns data as targets, 
RNN's easily overfit the training data
while failing to improve performance on in-sample validation.
\textbf{One key benefit of our approach}
is that by doing \emph{multi-task learning},
predicting all $16$ future fundamentals,
we provide the model with considerable training signal and may thus be less susceptible to overfitting.

The price movement of stocks is extremely noisy \cite{shiller1980stock} and so, suspecting that the relationships among fundamental data may have a larger signal to noise ratio than the relationship between fundamentals and price, we set up the problem thusly:
For MLPs, at each month $t$,
given features for $5$ months spaced $1$ year apart 
($t-48$, $t-36$, $t-24$, $t-12$), predict the fundamental data at time $t+12$. 
For RNNs, the setup is identical but with the small modification that for  each input in the sequence, we predict the corresponding $12$ month lookahead data.

We evaluated two classes of deep neural networks: MLPs and RNNs. For each of these, we tune hyperparameters
on the in-sample period. 
We then evaluated the resulting model on the out-of-sample period. 
For both MLPs and RNNs, we consider architectures evaluated with $1$, $2$, and $4$ layers with $64$, $128$, $256$, $512$ or $1024$ nodes. We also evaluate the use of dropout both on the inputs and between hidden layers. For MLPs we use ReLU activations and apply batch normalization between layers. For RNNs we test both GRU and LSTM cells with layer normalization. 
We also searched over various optimizers (SGD, AdaGrad, AdaDelta), settling on AdaDelta. We also applied L2-norm clipping on RNNs to prevent exploding gradients.
Our optimization objective is to minimize square loss.

To account for the fact that we care more about our prediction of EBIT over the other fundamental values, we up-weight it in the loss (introducing a hyperparameter $\alpha_1$).
For RNNs, because we care primarily about the accuracy of the prediction at the final time step (of 5), 
we upweight the loss at the final time step by hyperparameter $\alpha_2$ (as in \cite{lipton2016learning}).
Some results from our hyperparameter search on in-sample data are displayed in  
Table 1. 
These hyperparameters resulted in MSE on in-sample validation data of $0.6141$ for and $0.6109$ for the MLP and RNN, respectively.

  .\\
\begin{wrapfigure}{r}{0.5\textwidth}
\vspace{20px}
 \begin{tabular}{l|cc}
  \toprule
  Hyperparameter &MLP& RNN\\
  \midrule
   {Hidden Units}& $1024$ & $64$  \\
   {Hidden Layers}& $2$ & $2$  \\
   {Input Dropout Keep Prob.}&  $1.0$ & $1.0$  \\
   {Hidden Dropout Keep Prob.}& $0.5$ & $1.0$  \\
   {Recurrent Dropout Keep Prob.}& n/a & $0.7$  \\
   {Max Gradient Norm}& $1.0$ & $1.0$  \\
   {$\alpha_1$ }& $0.75$ & $0.5$  \\
   {$\alpha_2$ }& n/a & $0.7$  \\
  \bottomrule
\end{tabular}    
\label{tab:hyperparams}
\captionof{table}{Final hyperparameters for MLP and RNN}
\vspace{-5px}
\end{wrapfigure}

\paragraph{Evaluation}
As a first step in evaluating the forecast produced by the neural networks, 
we compare the MSE of the predicted fundamental on out-of-sample data with a naive prediction where predicted fundamentals at time $t$ is assumed to be the same as the fundamentals at $t-12$. 
To compare the practical utility of traditional factor models vs lookahead factor models
we employ an industry grade investment simulator.
The simulator evaluates hypothetical stock portfolios 
constructed on out-of-sample data. 
Simulated investment returns reflect how an investor might have performed 
had they invested in the past according to given strategy.

The simulation results reflect assets-under-management at the start of each month that, when adjusted by the S\&P 500 Index Price to January 2010, are equal to \$100 million. 
We construct portfolios by ranking all stocks according to the factor EBIT/EV in each month and investing equal amounts of capital into the top $50$ stocks 
holding each stock for one-year. 
When a stock falls out of the top 50 after one year, 
it is sold with proceeds reinvested 
in another highly ranked stock 
that is not currently in the simulated portfolio. 
We limit the number of shares of a security bought or sold in a month
to no more than $10\%$ of the monthly volume for a security. 
Simulated prices for stock purchases and sales are based 
on the volume-weighted daily closing price of the security 
during the first $10$ trading days of each month.
If a stock paid a dividend during the period it was held, 
the dividend was credited to the simulated fund 
in proportion to the shares held.
Transaction costs are factored in as \$0.01 per share, 
plus an additional slippage factor 
that increases as a square of the simulation’s volume participation in a security. Specifically, if participating at the maximum $10\%$ of monthly volume, 
the simulation buys at $1\%$ more than the average market price 
and sells at $1\%$ less than the average market price. 
Slippage accounts for transaction friction, such as bid/ask spreads, that exists in real life trading.

\textbf{Our results demonstrate} a clear advantage for the lookahead factor model.
In nearly all months, however turbulent the market,
neural networks outperform the naive predictor (that fundamentals remains unchanged) (Figure \ref{fig:out-of-sample-mse-mlp}).
Simulated portfolios lookahead factor strategies  
with MLP and RNN perform similarly, 
both beating traditional factor models (Table \ref{tab:simulation-results}).

\vspace{-5px}
\section{Discussion}
\label{sec:discussion}
In this paper we demonstrate a new approach for automated stock market prediction based on time series analysis.
Rather than predicting price directly, 
predict future fundamental data from a trailing window of values.
Retrospective analysis with an oracle motivates the approach,
demonstrating the superiority of LFM over standard factor approaches. 
In future work we will thoroughly investigate the relative advantages of LFMs vs directly predicting price. We also plan to investigate the effects of the sampling window, input length, and lookahead distance.

\bibliographystyle{plainnat}
\bibliography{main}

\end{document}